\documentclass[10pt,twocolumn,letterpaper]{article}

\usepackage{iccv}
\usepackage{times}
\usepackage{epsfig}
\usepackage{graphicx}
\usepackage{amsmath}
\usepackage{amssymb}
\usepackage[cjk]{kotex}
\usepackage{booktabs}
\usepackage{colortbl}
\usepackage{pifont}
\usepackage{pdfrender}
\usepackage{makecell}
\usepackage{tabularx}
\usepackage{colortbl}
\usepackage{marvosym}
\usepackage{hyperref}
\usepackage{xcolor}
\usepackage{tabularray}
\usepackage{xurl}
\usepackage{subcaption}
\usepackage{tikz}
\usepackage{multirow}
\hypersetup{
    colorlinks=False,
    linkcolor=black,
    filecolor=black,      
    urlcolor=black,
}
\usepackage{floatrow}
\usepackage{tablefootnote}
\usepackage[flushleft]{threeparttable}

\usepackage[T1]{fontenc}
\newcommand\ours{RAL\xspace}
\newcommand{\xmark}{\textcolor{black}{\ding{55}}}

\iccvfinalcopy 

\def\iccvPaperID{****} 

\ificcvfinal\fi

\begin{document}
\title{Robust Asymmetric Loss for Multi-Label Long-Tailed Learning}

\author{
  Wongi Park$^1$\thanks{Equal contribution.} , Inhyuk Park$^2$\footnotemark[1] , Sungeun Kim$^2$, and Jongbin Ryu$^1$$^2$\thanks{Corresponding author.}\footnotemark[2] \\
  {$^1$ Department of Software and Computer Engineering, Ajou University}  \\
		{$^2$ Department of Artificial Intelligence, Ajou University} \\
  \texttt{\{psboys, inhyuk, kimsungeun, jongbinryu\}@ajou.ac.kr}
}

\maketitle
\ificcvfinal\thispagestyle{empty}\fi

\begin{abstract}
In real medical data, training samples typically show long-tailed distributions with multiple labels. Class distribution of the medical data has a long-tailed shape, in which the incidence of different diseases is quite varied, and at the same time, it is not unusual for images taken from symptomatic patients to be multi-label diseases.
Therefore, in this paper, we concurrently address these two issues by putting forth a robust asymmetric loss on the polynomial function.
Since our loss tackles both long-tailed and multi-label classification problems simultaneously, it leads to a complex design of the loss function with a large number of hyper-parameters. 
%
Although a model can be highly fine-tuned due to a large number of hyper-parameters, it is difficult to optimize all hyper-parameters at the same time, and there might be a risk of overfitting a model. 
Therefore, we regularize the loss function using the Hill loss approach, which is beneficial to be less sensitive against the numerous hyper-parameters so that it reduces the risk of overfitting the model.
For this reason, the proposed loss is a generic method that can be applied to most medical image classification tasks and does not make the training process more time-consuming. 
We demonstrate that the proposed robust asymmetric loss performs favorably against the long-tailed with multi-label medical image classification in addition to the various long-tailed single-label datasets.
Notably, our method achieves Top-5 results on the CXR-LT dataset of the ICCV CVAMD 2023 competition. We opensource our implementation of the robust asymmetric loss in the public repository: \href{https://github.com/kalelpark/RAL}{https://github.com/kalelpark/RAL}.
\end{abstract}


\begin{figure}[!t]
\centering
\includegraphics[width=\textwidth]{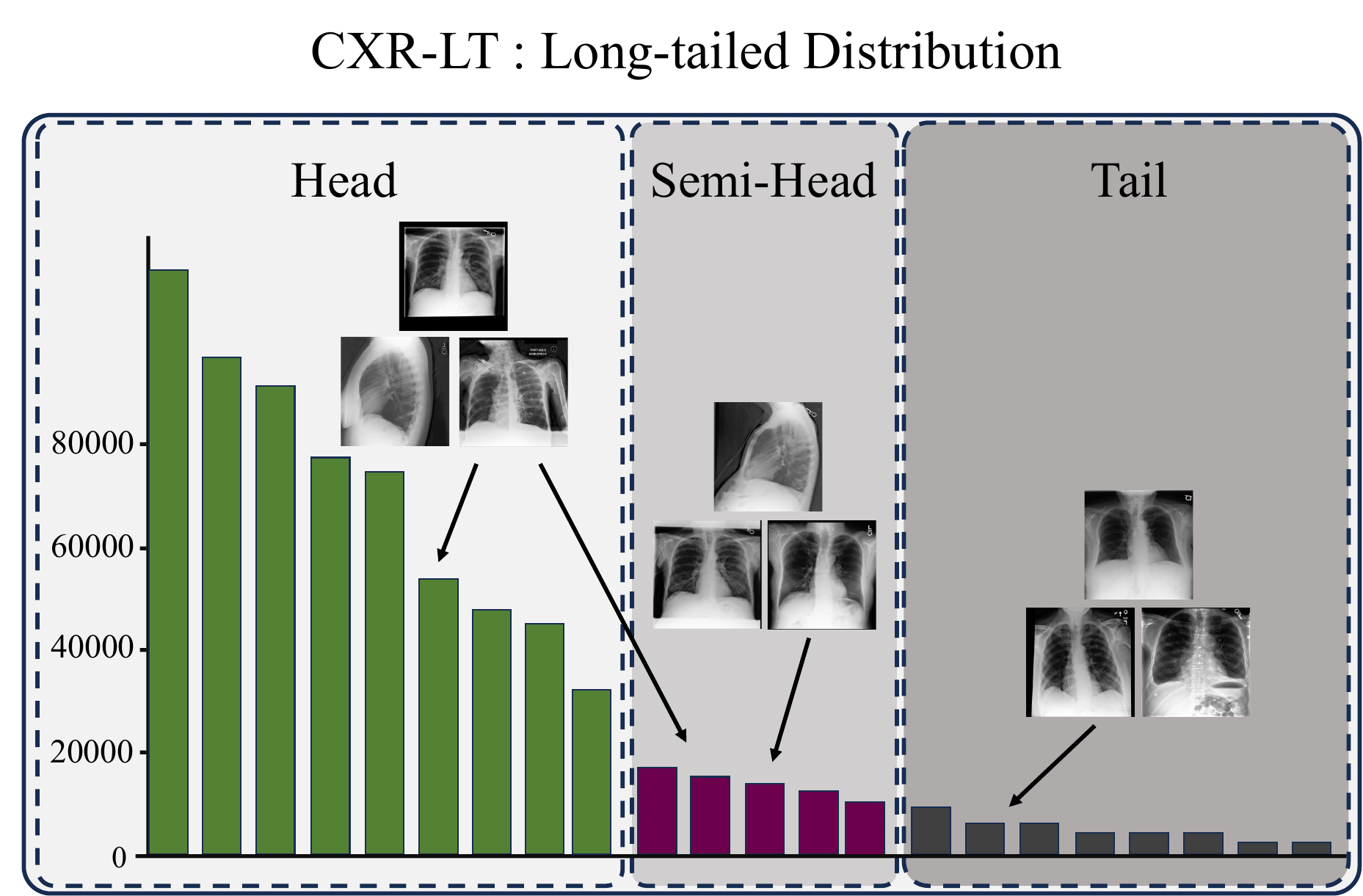}
\caption{Label distribution of the CXR-LT Dataset\cite{CXRLT}. 
Typical radiological images have a lengthy tail distribution since there are few positive samples for certain classes. Further single radiological image contains multiple classes in most cases. These long-tailed and multi-label data is common but critical issues in real-world medical image recognition task.
}

\label{fig:dataset}
\end{figure}

\begin{figure*}[!t]
\centering
\includegraphics[width=0.9\textwidth]{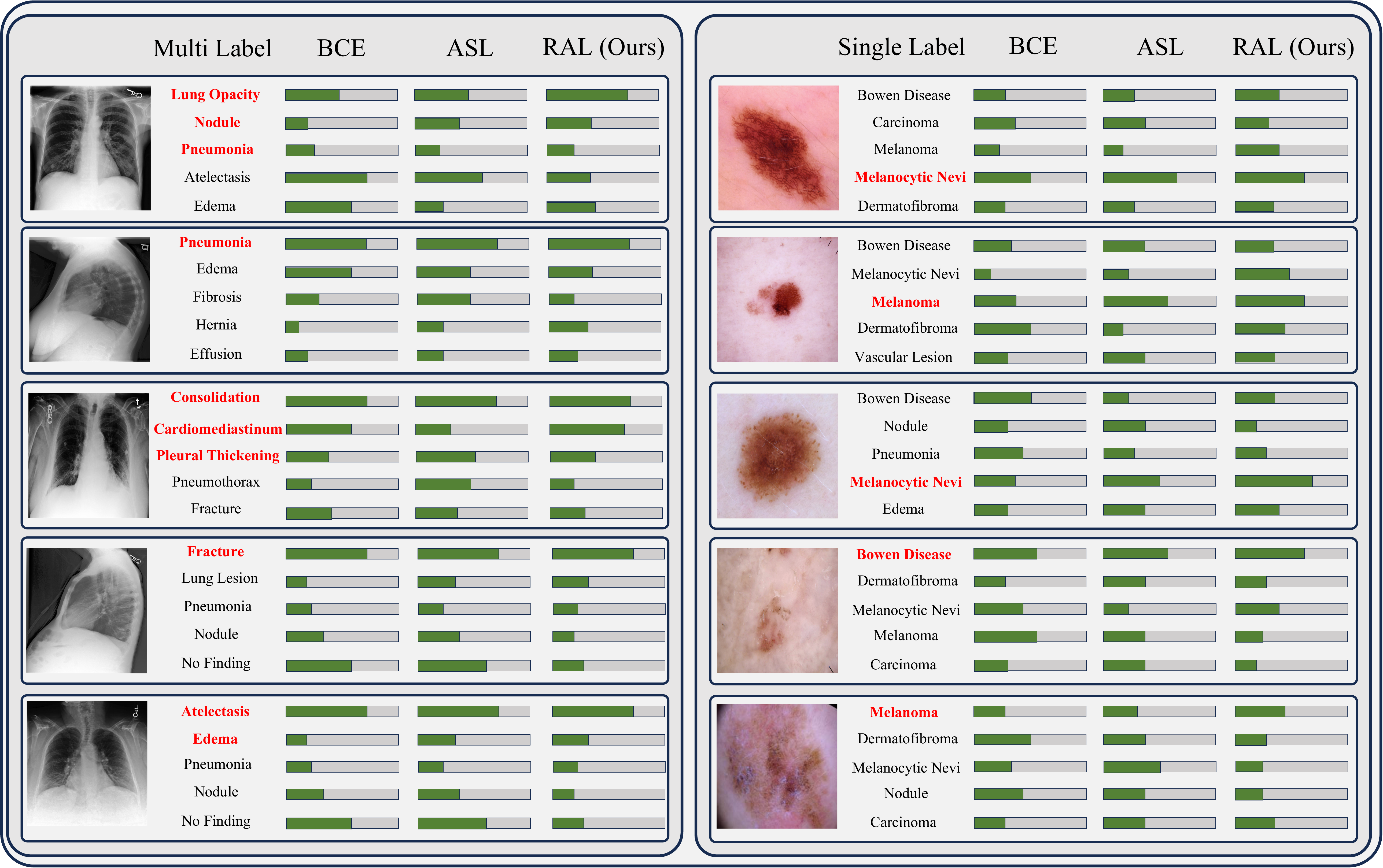}
\caption{Examples of estimated probabilities of the BCE, ASL, and our RAL from the CXR-LT\cite{CXRLT} and ISIC2018\cite{ISIC2018} datasets. 
Example result of our proposed methods, BCE and ASL\cite{ben2020asymmetric} from CXR-LT Dataset\cite{CXRLT} and ISIC2018 Dataset\cite{ISIC2018}. 
The color red denotes the positive labels. The model trained by BCE loss has a tendency to overfit the single label. On the other hand, models trained by ours exhibit higher probabilities for multiple labels, implying that ours are more reliable for the long-tailed multi-label classification task.
}

\label{fig:FG_SSL}
\end{figure*}
\section{Introduction}
Multi-label classification, which predicts more than one label from a single image, has received lots of interest in recent years.
Especially in the field of medical image recognition, several studies \cite{agu2021anaxnet, maksoud2019medical, zhou2022long, sadafi2019multiclass} have been conducted to tackle the problem of the coexistence of multiple symptoms in a single radiology image.
However, these multi-label classification studies have overlooked another critical issue: the long-tailed distribution of medical data. \cite{chen2019deep, bhusal2022multi, zhou2022long, holste2022long} 
That is the long-tailed distribution problem. In general, multi-label data, the more classes there are used, the more long-tailed the distribution. 
In this case, for classes with fewer labels (tail labels), the performance of the model will drop significantly, and the model will be biased to the data of head labels with more training data. Therefore, it is challenging to generalize the learned model in practice.
To solve this imbalance data distribution, there have been studies that re-sample\cite{ando2017deep, pouyanfar2018dynamic, shen2016relay} or re-weight\cite{yang2020rethinking, huang2019deep, dong2018imbalanced} the data to make the model learn more from the tail labels, but these studies have not been addressed the long-tailed problem in a multi-label classification environment.

Another issue is that many studies\cite{xiao2023delving, ge2020improving, wu2020comprehensive} exploit additional resources to tackle the long-tailed distribution and multi-label classification problems. 
Using larger models or increasing computational complexity can help such problems, but there might be limitations in that their high budget reduces the practical applicability.
Therefore, in this paper, we propose a robust asymmetric loss that does not require additional resources to learn the multi-label long-tailed medical data. 
The proposed loss is based on asymmetric weighting, which ensures that the importance of the negative sample's loss is regarded differently from that of positive samples so that even hard negative samples can be robustly learned.
Compared to the existing cross-entropy, focal loss\cite{lin2017focal}, asymmetric loss\cite{ben2020asymmetric}, and balanced loss\cite{cui2019class}, the proposed robust asymmetric focal loss effectively learns long-tailed multi-label data reliably while being less sensitive to hyper-parameters. 
Specifically, we re-weight the negative samples adopting the hill loss\cite{zhang2021simple} so that ours performs favorably against the hard negative samples while being robust to the settings on the variety of hyper-parameters.
To this end, we expand the asymmetric loss using a Taylor series-based approach\cite{huang2023asymmetric} to account for the negative loss. The Taylor series ensures that negative samples below a certain threshold are not used for training so that the stable gradient can be passed to the deep neural networks.

\begin{figure*}[!t]
\centering
\includegraphics[width=0.98\textwidth]{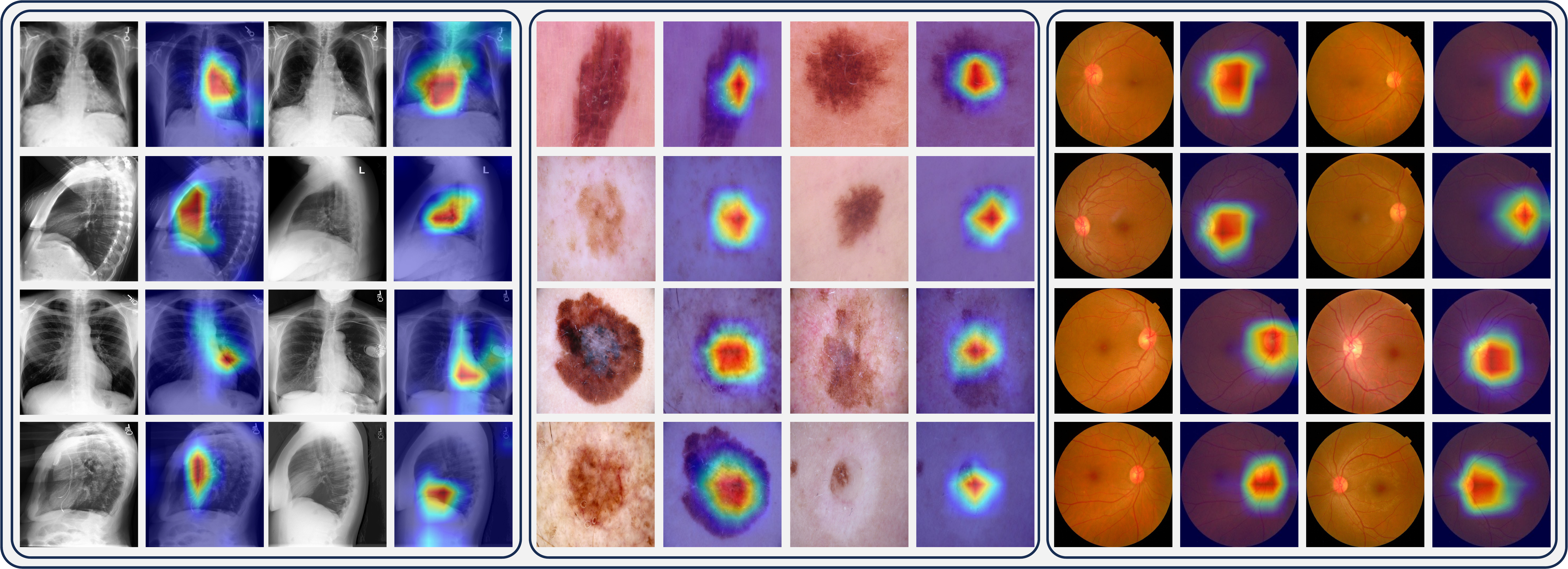}
\caption{Grad-CAM visualization \cite{selvaraju2017grad} from the model trained by the proposed RAL.}
\label{fig:med_cam}
\end{figure*}

We evaluate the proposed method on CXR-LT, a long-tailed multi-label classification medical dataset, and demonstrate that ours improve the performance of the classification task over existing methods. Notably, our method achieved 0.351 mAP, which is within the Top-5 of the final ranking in the ICCV CVAMD 2023 competition. 
Furthermore, we evaluate that our robust asymmetric loss works well on long-tailed distributions, even on single-label medical datasets. For this evaluation, we utilize the ISIC2018 and APTOS2019 datasets and show that our method achieves considerably better performance compared to the existing methods.

We present the contributions of this paper as follows.
\begin{itemize}
\item
We propose robust asymmetric loss, which is effective for long-tailed multi-label classification. The proposed loss can be finely tuned but is not sensitive to hyper-parameter settings.
\item 
We improve the performance of long-tailed multi-label classification without additional training data, model parameters, and computational budget.
\item
We achieved Top-5 results in the CVAMD2023 competition on the long-tailed multi-label CXR-LT dataset. In addition, we confirm that the proposed method works well on single-label medical image classification as well as the multi-label dataset.
\end{itemize}


\section{Related Work}
\textbf{Multi-Label} classification has been extensively studied to predict more than one class label\cite{liu2021emerging, han2023survey}. Recently, to understand the correlation between multiple labels, several studies have introduced network architectures that enable the model to predict the inherent relation between features and corresponding labels. 
Most of the studies have used Graph Convolution Network(GCN)\cite{chen2019multi, ye2020attention, chen2021learning} that learn label's feature relation in the graph structure. Subsequently, semantic representation in images using attention mechanism has been researched in many studies\cite{kovtun2023label, wang2022cross}. 
On the other hand, there have been training algorithms to tackle the multi-label classification, such as investigation on the weight re-weighting and class frequency\cite{cui2019class, park2021influence, samuel2021distributional}. More recently, asymmetric loss\cite{ben2020asymmetric,huang2023asymmetric} has been introduced to optimize the imbalanced positive and negative losses. 

\textbf{Long-tailed} distribution has been regarded as a practical problem for real-world machine learning applications. Studies that address the problem of the long-tailed distribution can be divided into two categories.
First, re-sampling methods\cite{zang2021fasa, ju2022flexible, galdran2021balanced} have been introduced to under-sample or over-sample the data according to its class distribution to construct the balanced training set. 
Re-weighting strategy assigns different weights to the samples to adjust the long-tailed distribution\cite{lin2017focal, cui2019class}. 
However, there is an ambiguity in applying the re-sampling methods to multi-label datasets. When a single image contains both head and tail class labels, it is difficult to determine whether the sample should be over- or under-sampled.
%
%
For this reason, re-sampling methods are hardly applicable to the multi-label classification so we encounter the problem that most gradients are computed from negative samples.
To address this issue, the second category, studies\cite{du2023adaptive, park2021influence, samuel2021distributional} for dealing with the long-tailed class distribution exploiting the loss functions, are being researched.
The focal loss\cite{lin2017focal} is the landmark method to tackle the long-tailed class distribution using the loss function.
%
Focal loss adds the modulating factors(\ie focusing and balance parameters) to the cross-entropy loss so that it can mitigate the long-tailed distribution by controlling such modulating factors. 
Then, to further tailor the loss function, asymmetric loss\cite{huang2023asymmetric, ridnik2021asymmetric}  was proposed, which determines the focusing parameters of negative and positive loss separately.
Recently, the loss function has been expanded to polynomial functions\cite{huang2023asymmetric} to use only several principle terms in computing the gradients.
On the other hand, the Hill loss\cite{zhang2021simple} was proposed to prevent the gradient from being too large in certain samples.
The Hill loss performed well for multi-label classification, but they did not apply their method to complex formulas unfolded as polynomial.
Therefore, in this paper, we apply the Hill term loss to the formula extended to polynomials to learn a long-tailed multi-label classification model more accurately and robustly.
\section{Method}
In this section, we introduce our robust asymmetric loss function. We first describe the existing long-tailed and multi-class losses as the background of our loss function. Then, we introduce the robust asymmetric loss function by adding the Hill loss term to the polynomial function. 


\begin{table*}[!t]
\centering
\begin{tblr}{
  width = \textwidth,
  colspec = {Q[206]Q[119]Q[131]Q[215]Q[119]Q[131]},
  cells = {c},
  cell{1}{1} = {r=2}{},
  cell{1}{2} = {c=2}{0.25\textwidth},
  cell{1}{4} = {r=2}{},
  cell{1}{5} = {c=2}{0.25\textwidth},
  hline{1,16} = {-}{0.08em},
  hline{2} = {2-3,5-6}{0.03em},
  hline{3} = {-}{0.05em},
  rowsep = 1mm
}
Label             & Positive  &             & Label              & Positive  &             \\
                  & \#Sample(K) & Portion(\%) &                    & \#Sample(K) & Portion(\%) \\
Atelectasis       & 67.6      & 10.6        & Mass               & 5.5       & 0.9         \\[-0.3em]
Calcification     & 4.3       & 0.7         & No Finding         & 41.8      & 6.6         \\[-0.3em]
Cardiomegaly      & 76.9      & 12.1        & Nodule             & 7.6       & 1.2         \\[-0.3em]
Consolidation     & 16.0      & 2.5         & Pleural Effusion   & 69.2      & 10.8        \\[-0.3em]
Edema             & 38.6      & 6.1         & Pleural Other      & 0.6       & 0.1         \\[-0.3em]
Emphysema         & 4.3       & 0.7         & Pleural Thickening & 3.3       & 0.5         \\[-0.3em]
Cardiomediastinum & 30.1      & 4.7         & Pneumomediastinum  & 0.7       & 0.1         \\[-0.3em]
Fibrosis          & 1.1       & 0.2         & Pneumonia          & 49.1      & 7.6         \\[-0.3em]
Fracture          & 11.9      & 1.9         & Pneumoperitoneum   & 0.5       & 0.1         \\[-0.3em]
Hernia            & 4.0       & 0.6         & Pneumothorax       & 14.9      & 2.4         \\[-0.3em]
Infiltration      & 10.2      & 1.6         & Emphysema          & 2.4       & 0.4         \\[-0.3em]
Lung Lesion       & 2.5       & 0.4         & Support Devices    & 89.1      & 14          \\[-0.3em]
Lung Opacity      & 79.9      & 12.6        & Tortuous Aorta     & 3.4       & 0.6  
\end{tblr}
\caption{Specification of the CXR-LT dataset. It shows that the samples are heavily distributed in a few classes while several classes have very few samples.}
\label{tab:descript_dataset}
\end{table*}

\subsection{Long-tailed and Multi-label Classification Loss} \label{loss_func}
Traditionally, multi-label classification tasks use the Binary Cross-Entropy (BCE) Loss as:
\begin{equation}
    \mathcal{L_{BCE}} = -\sum_{i=1}^{K}\left(y_{i}L_{i}^{+} + (1 - y_{i})L_{i}^{-} \right)
\end{equation}
\begin{equation}
\left\{\begin{array}{l}
\mathcal{L}^{+}=\log (\hat{y}) \\
\mathcal{L}^{-}=\log (1-\hat{y}) 
\end{array}\right.,
\end{equation}
\noindent where $L_{i}^{+}$ and $L_{i}^{-}$ are positive and negative sample losses and $y$ and $\hat{y}$ denote the ground-truth and estimated probability for the class labels. 

However, since this BCE function computes the same weights for all class samples in training data with the long-tailed distribution, it excessively focuses on learning the head classes with a large number of training samples. This problem is addressed by the focal loss\cite{lin2017focal} $\mathcal{L}_{Focal}$ with balancing the positive and negative losses as:
\begin{equation}
\left\{\begin{array}{l}
\mathcal{L}_{Focal}^{+}=\alpha_{+}(1- \hat{y})^{\gamma}\log(\hat{y}) \\
\mathcal{L}_{Focal}^{-}=\alpha_{-}\hat{y}^{\gamma}\log(1-\hat{y}) 
\end{array}\right.,
\end{equation}
where $\alpha_{+}$ and $\alpha_{-}$ represent the balancing parameter and $\gamma$ denotes the focusing parameter that is the key hyper-parameters of the focal loss function.
Controlling the hyper-parameters, the focal loss can balance the head- and tail-class samples. However, this focal loss has a weakness in that positive and negative losses share the same focusing parameter $\gamma$.
Therefore asymmetric weighting approach\cite{ridnik2021asymmetric} for the loss function alleviates this problem by assigning different focusing parameters as:
\begin{equation}
\left\{\begin{array}{l}
\mathcal{L}_{ASL}^{+}=(1- \hat{y})^{\gamma+}\log(\hat{y}) \\
\mathcal{L}_{ASL}^{-}=\hat{y}_{\tau}^{\gamma-}\log(1-\hat{y}_{\tau}) 
\end{array}\right.
\end{equation}
\[
\hat{y}_{\tau} = max(\hat{y} - \tau, 0),
\]
where $\gamma^{+}$ and $\gamma^{-}$ are the positive and negative focusing parameters and $y_{\tau}$ denotes the rectified probability thresholded by $\tau$.
This ASymmetric Loss (ASL) is efficient for optimizing the training of positive and negative samples separately and able to mitigate the gradient vanishing problem due to too small a value of $\hat{y}$ in the negative loss.

\subsection{Robust Asymmetric Loss} \label{asymmetricloss}
Asymmetric loss can be expanded to the polynomial equation using the Taylor series\cite{linnainmaa1976taylor} based method\cite{huang2023asymmetric}.
In a polynomial equation, using several principle low-order terms can improve the performance of multi-label classification tasks.
This is because the higher-order terms in the polynomial form can be regarded as noise or redundant, so using only a few low-order terms is effective.
Therefore, the asymmetric polynomial loss is formulated as follows: 
\begin{equation}
\left\{\begin{array}{l}
\mathcal{L}_{APL}^{+}=y \sum_{m=1}^{M} \alpha_{m}(1-\hat{y})^{m+\gamma^{+}} \\
\mathcal{L}_{APL}^{-}=(1-y) \sum_{n=1}^{N} \beta_{n} \hat{y}_{\tau}^{n+\gamma^{-}}
\end{array}\right.,
\end{equation}
where $M$ and $N$ are parameters that determine the number of low-order terms to be used in the positive and negative losses, and $\alpha_{m}$ and $\beta_{n}$ stand for the balance parameter of each term in the positive and negative losses. 
This Asymmetric Polynomial Loss (APL)\cite{huang2023asymmetric} has the advantage of controlling the positive and negative losses on a term-by-term basis, but it also has the significant drawback of requiring a large number of hyper-parameters to be configured by the user. Optimizing such a large number of hyper-parameters can be a time-consuming process and often leads to overfitting the models.

To be less sensitive to optimizing the numerous hyper-parameters, we introduce robust asymmetric loss. It is noticeable that, especially in multi-label data, the number of negative samples is much larger than that of positives, so making the negative loss less sensitive is the most decisive factor in the long-tailed multi-label classification task.
Therefore, we adopt the Hill loss\cite{zhang2021simple} so that we prevent an excessively large gradient of the negative loss in the learning process. Adding the Hill loss term to APL, we define our Robust Asymmetric Loss (RAL) as: 
\begin{equation}
\left\{\begin{array}{l}
\mathcal{L}_{RAL}^{+}=y \sum_{m=1}^{M} \alpha_{m}(1-\hat{y})^{m+\gamma^{+}} \\
\mathcal{L}_{RAL}^{-}=\psi(\hat{y}) \cdot (1-y) \sum_{n=1}^{N} \beta_{n} \hat{y}_{\tau}^{n+\gamma^{-}}
\end{array}\right.
\label{eq:RAL}
\end{equation}
\[
\psi(\hat{y}) = \lambda - \hat{y},
\]
where $\psi$ denotes the Hill loss term and $\lambda$ is set to $1.5$ value.
Our RAL is robust to the change of numerous hyper-parameters due to the less sensitive negative loss in the training process.
In the negative loss, when $\hat{y}$ is close to 0, that is, the estimated probability of the training data is close to the correct negative answer, the gradient value is already small. Therefore, in this case, the hyper-parameter is not sensitive. On the other hand, when the estimated probability is around 0, which is a hard negative sample, the gradient value is too large, making the network training sensitive to the hyper-parameter settings. Our RAL loss regularizes the gradient of these hard negative samples to make them less sensitive to hyper-parameters.
As we expand the asymmetric loss to polynomial form, there is an unavoidable problem of setting too many hyper-parameter, so we propose RAL with Hill loss term to alleviate such a problem.

\begin{table}[!h]
\centering
\renewcommand{\arraystretch}{1.1}
\setlength{\tabcolsep}{5pt}
    \begin{tabular}{cccc}
    \toprule
    Dataset                     & classes & Samples & Imbalance Ratio \\ \midrule
    CXR-LT                     & 26 & 377$,$110 & 142 \\
    APTOS2019                     & 7 & 10$,$015 & 58 \\
    ISIC2018                     & 5 & 3$,$662 & 10 \\ \bottomrule
    \end{tabular}
    \caption{The details of long-tailed medical datasets.}
    \label{tab:detail_set}
\end{table}

\begin{table}[!bht]
\centering
\resizebox{\columnwidth}{!}{
\begin{tblr}{
  cells = {c},
  cell{2}{1} = {r=2}{},
  cell{4}{1} = {r=2}{},
  cell{6}{1} = {r=2}{},
  cell{8}{1} = {r=2}{},
  hline{1-2} = {-}{},
  hline{10} = {-}{0.08em},
  rowsep = 1mm
}
Method                   & Image size & mAP            & mAUC           & mF1            \\
CE                       & 224       & 0.301          & 0.808          & 0.218          \\
                         & 384       & 0.314          & 0.813          & 0.227          \\
Focal loss\cite{lin2017focal}               & 224       & 0.304          & 0.807          & 0.231          \\
                         & 384       & 0.295          & 0.803          & 0.224          \\
ASL\cite{ben2020asymmetric}                 & 224       & 0.307          & 0.808          & 0.225          \\
                         & 384       & 0.317          & 0.811          & 0.237 \\
RAL (Ours) & 224       & 0.314          & 0.815          & 0.225          \\
                         & 384       & 0.323 & 0.817 & 0.233          
\end{tblr}
 }
\caption{Experimental comparison on the loss functions with ours.}
\label{tab:sota_table}
\end{table}

\section{Experiments Setup} \label{experiments} 
\subsection{Dataset and Metrics} 
\noindent \textbf{CXR-LT.} The 377,110 CXRs in the ICCV CVAMD 2023 Dataset(CXR-LT) dataset, which is included in the competition, have at least one label in 26 clinical findings.
The class labels consist of the "No Finding" class and 12 new disease labels introduced from mimic-cxr-jpg, which cover chest X-rays(CXR). The detailed specification of the CXR-LT dataset can be found in Table \ref{tab:descript_dataset}.
%
We randomly divide the image sets into a test and training set with a ratio of 8:2 for the CXR-LT dataset. \\
\textbf{ISIC2018 and APTOS2019. } 
The ISISC2018 dataset has $10,015$ skin images with 7 lesion classes, and the APTOS dataset includes $3,662$ diabetic retinopathy images with 5 disease classes.
For these APTOS2019 and ISIC2018 datasets, we follow the same protocol of the previous study\cite{marrakchi2021fighting}.  \\
\textbf{Metric.} To evaluate our methods for the CXR-LT dataset, we use three metrics such as mean Average Precision(mAP), mean Area Under Curve(mAUC), and F1-Score considering the multi-label dataset. For APTOS2019 and ISIC2018, which is the single-label dataset, we use two metrics as Accuracy and F1-Score. The details of these three datasets are specified in Table \ref{tab:detail_set}. The imbalance ratio for measuring the significance of the long-tailed distribution is denoted as $N_{max} / N_{min}$, where $N$ is the number of each class sample in each class.

\begin{table*}[!t]
\centering
\renewcommand{\arraystretch}{1.25}
\setlength{\tabcolsep}{8.9pt}
    \begin{tabular}{cccccccc}
    \toprule
    Label                      & BCE & APL & RAL (Ours)                & Label                & BCE & APL & RAL (Ours) \\ \hline
    Atelectasis                &  0.578  &  0.599   &  0.610    & Mass                   &  0.159  &  0.200  &  0.222    \\
    Calcification              &  0.120   &  0.137   &  0.151    & No Finding             &  0.445   &  0.477   &  0.479    \\
    Cardiomegaly               &  0.626   &  0.633   &  0.648    & Nodule                 &  0.148   &  0.205   &  0.234    \\
    Consolidation              &  0.203  &  0.209    &  0.224    & Pleural Effusion       &  0.801  &  0.813   &  0.821    \\
    Edema                      &  0.527  &  0.552    &  0.562    & Pleural Other          &  0.015   & 0.048  &  0.059    \\
    Emphysema                  &  0.258   &  0.313   &  0.334    & Pleural Thickening     &  0.065   &  0.094  &  0.111    \\
    Cardiomediastinum          &  0.155   &  0.166   &  0.173    & Pneumomediastinum      &  0.103   &  0.138   &  0.203    \\
    Fibrosis                   &  0.099   &  0.121   &  0.131    & Pneumonia              &  0.289  &  0.306   &  0.167    \\
    Fracture                   &  0.175   &  0.223   &  0.270    & Pneumoperitoneum       &  0.134   &  0.143   &  0.524    \\
    Hernia                     &  0.483   &  0.509   &  0.560    & Pneumothorax           &  0.394   & 0.478  &  0.483    \\
    Infiltration               &  0.058   &  0.061   &  0.075    & Emphysema              & 0.391   &  0.459   &  0.544   \\
    Lung Lesion                &  0.054   &  0.059   &  0.079    & Support Devices        & 0.892  &  0.906 &  0.913    \\
    Lung Opacity               &   0.579  &  0.601   &  0.613    & Tortuous Aorta         &  0.055   &  0.052  &  0.056    \\ \bottomrule
    \end{tabular}
    \caption{Experimental results of the development phase of the CVAMD 2023 competition. Our RAL works well on most cases compared to the other loss functions.}
    \label{tab:dev_table}
\end{table*}

\subsection{Implementation details.}\label{Imp_details}  
We use the ConvNeXT-B\cite{liu2022convnet} as the backbone for the proposed loss. 
We resize the input images as $384 \times 384$ and exploit the data augmentation schemes following the previous\cite{azizi2021big, chen2019multi}.
We train our networks using the Adam optimizer with 0.9 momentum and 0.001 weight decay. The batch size is 256, and the initial learning rate is set to $1e-4$. Our networks are trained on PyTorch version 1.11.0 with RTX A5000 GPUs.

\section{Experimental results}
In this section, we show the experimental results to validate the effectiveness of \ours.
%
We first compare the proposed RAL with previous state-of-the-art loss functions such as focal loss, LDAM, and ASL.
We then dissect the proposed loss function into its component level to demonstrate its robustness.
%
In this experiment, \ours performs well consistently for variations of numerous hyper-parameters. We also show that the proposed \ours works favorably on both multi- and single-label long-tailed medical image classification tasks. Further, we validate that ours is robust to several noisy conditions.

        

\begin{table}[!h]
\centering
\setlength{\tabcolsep}{1.0pt}
\renewcommand{\arraystretch}{0.95}
\begin{tabular}{c*{4}{c}}
    \hline
     \multirow{ 2}{*}{method} & \multicolumn{2}{c}{ISIC2018}     &          \multicolumn{2}{c}{APTOS2019} \\
            & Accuracy & F1-score & Accuracy & F1-score\\
    \hline
    CE & 0.850 & 0.716 & 0.812 & 0.608 \\
    Focal loss\cite{lin2017focal} & 0.861 & 0.735 & 0.815 & 0.629 \\
    LDAM\cite{cao2019learning} & 0.849 & 0.728 & 0.813 & 0.620 \\
    ASL$^\dagger$ \cite{ben2020asymmetric} & 0.854 & 0.734 & 0.820 & 0.660 \\
    RAL (Ours) & 0.852 & 0.740 & 0.826 & 0.673 \\
    \hline
    \end{tabular}
    \caption{Experimental results of the ICIS2018 and APTOS2019 datasets. $\dagger$ denotes the result from our implementation with the official code: \href{https://github.com/Alibaba-MIIL/ASL}{https://github.com/Alibaba-MIIL/ASL}.}
\label{tab:single_label}
\end{table}

\subsection{Comparison on Loss Functions} \label{experimentalresults}
In this subsection, we compare our RAL with other loss functions on three datasets such as ISIC2018, APTOS2019, and CXR-LT.
In this part, we compare the proposed RAl with other methods on three datasets: ISIC2018, APTOS2019, and CXR-LT. 
Table\ref{tab:sota_table} and \ref{tab:dev_table} show the result submitted to the CVAMD 2023 competition site using our RAL at the development phase. In this result, our RAL achieves competitive performance compared to others. Ours performs well on most classes consistently in Table \ref{tab:dev_table}.
Further, our proposed RAL works well on single-label long-tailed datasets, such as ISIC2018 and APTOS2019. It outperforms the other methods in such datasets in Table \ref{tab:single_lable}.
These findings of the experimental results highlight the competitiveness of our RAL in diverse long-tailed medical image classification datasets.

\subsection{Ablation Study}
For a more in-depth analysis of the proposed method, we broke \ours into three components in our ablation study.
%
The three components are focal, asymmetric, and Hill loss where we apply the polynomial expansion to the Hill loss.
%
All results of this ablation study are taken using the ConvNeXt-B model with $384 \times 384$ image size in Table \ref{tab:component_table}.
Through this ablation study, we demonstrate that each component of our RAL is effective for the long-tailed multi-label classification task.
\begin{table}[!h]
    \centering
    \renewcommand{\arraystretch}{1.1}
    \begin{tabular}{cccccc}
        \toprule    
            Focal Loss & Asymmetric &  Hill &   mAP & mAUC   \\ \hline
            \checkmark & \xmark           & \xmark        & 0.295 & 0.803   \\ 
            \checkmark & \checkmark & \xmark        & 0.307 & 0.815   \\
            \textbf{\checkmark} & \textbf{\checkmark} & \textbf{\checkmark} & 0.323 & 0.817  \\ 
        \bottomrule
    \end{tabular}
    \caption{Experimental result of the ablation study of the proposed RAL.}
    \label{tab:component_table}
\end{table}




%
%

\begin{table*}[!t]
\centering
\resizebox{\textwidth}{!}{
\begin{tblr}{
  width = \textwidth,
  colspec = {Q[229]Q[77]Q[77]Q[77]Q[242]Q[77]Q[77]Q[77]},
  cells = {c},
  hline{1,16} = {-}{1pt},
  hline{2} = {2-4,6-8}{},
  hline{3} = {1-8}{0.6pt},
  rowsep = 1.2mm
}
Image size        & 1024       & 1512      & 2048       & Image size        & 1024       & 1512      & 2048       \\
Label             & \# 1       & \# 2  & \# 3  &    Label           & \# 1       & \# 2  & \# 3  \\
Atelectasis       & 0.607      & 0.576 & 0.574 & Mass               & 0.206      & 0.127 & 0.128 \\[-0.3em]
Calcification     & 0.142      & 0.111 & 0.115 & No Finding         & 0.478      & 0.452 & 0.452 \\[-0.3em]
Cardiomegaly      & 0.648      & 0.610 & 0.610 & Nodule             & 0.200      & 0.184 & 0.182 \\[-0.3em]
Consolidation     & 0.218      & 0.182 & 0.182 & Pleural Effusion   & 0.831      & 0.814 & 0.815 \\[-0.3em]
Edema             & 0.555      & 0.518 & 0.517 & Pleural Other      & 0.039      & 0.032 & 0.033 \\[-0.3em]
Emphysema         & 0.193      & 0.174 & 0.176 & Pleural Thickening & 0.109      & 0.082 & 0.083 \\[-0.3em]
Cardiomediastinum & 0.184      & 0.164 & 0.164 & Pneumomediastinum  & 0.338      & 0.305 & 0.313 \\[-0.3em]
Fibrosis          & 0.153      & 0.119 & 0.122 & Pneumonia          & 0.309      & 0.284 & 0.286 \\[-0.3em]
Fracture          & 0.289      & 0.234 & 0.237 & Pneumoperitoneum   & 0.282      & 0.239 & 0.230 \\[-0.3em]
Hernia            & 0.550      & 0.397 & 0.402 & Pneumothorax       & 0.552      & 0.507 & 0.507 \\[-0.3em]
Infiltration      & 0.060      & 0.056 & 0.055 & Emphysema          & 0.560      & 0.545 & 0.544 \\[-0.3em]
Lung Lesion       & 0.038      & 0.028 & 0.029 & Support Devices    & 0.913      & 0.896 & 0.896 \\[-0.3em]
Lung Opacity      & 0.596      & 0.556 & 0.556 & Tortuous Aorta     & 0.060      & 0.049 & 0.049 \\
\end{tblr}
}
\caption{Experimental result of test phase of the CVAMD 2023 competition. It shows that image size of $1024 \times 1024$ achieves the best result. We assume that this result is because the resolution we used for training is $1024 \times 1024$, which is not very large.}
\end{table*}

\begin{figure*}[!t]
  \centering
  \begin{subfigure}{0.325\textwidth}
    \includegraphics[width=\linewidth]{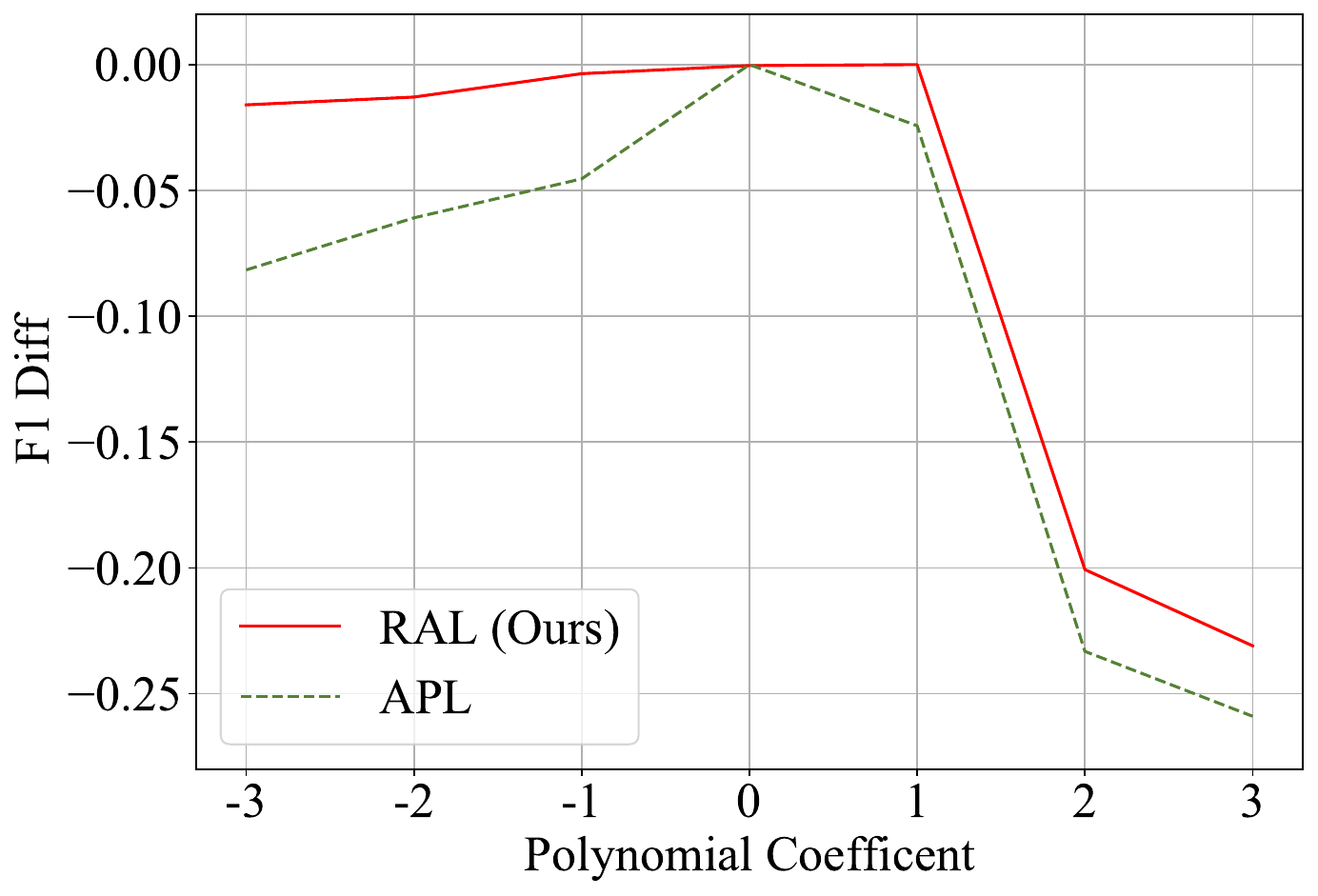}
    \caption{CXR-LT}
    \label{fig:cxr-lt-pow}
  \end{subfigure}\hfill
  \begin{subfigure}{0.33\textwidth}
    \includegraphics[width=\linewidth]{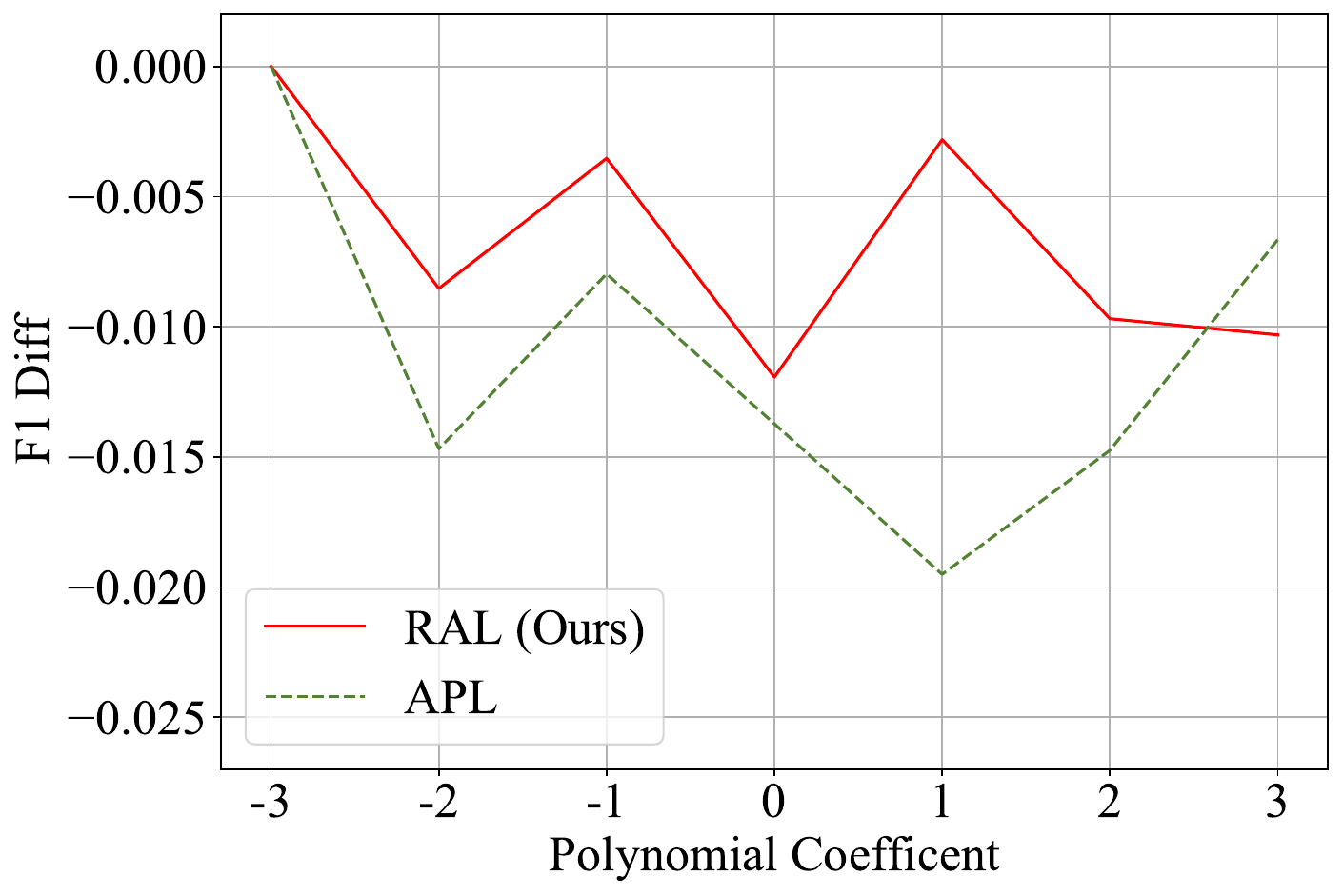}
    \caption{ISIC2018}
    \label{fig:isic-pow}
  \end{subfigure}\hfill
  \begin{subfigure}{0.328\textwidth}
    \includegraphics[width=\linewidth]{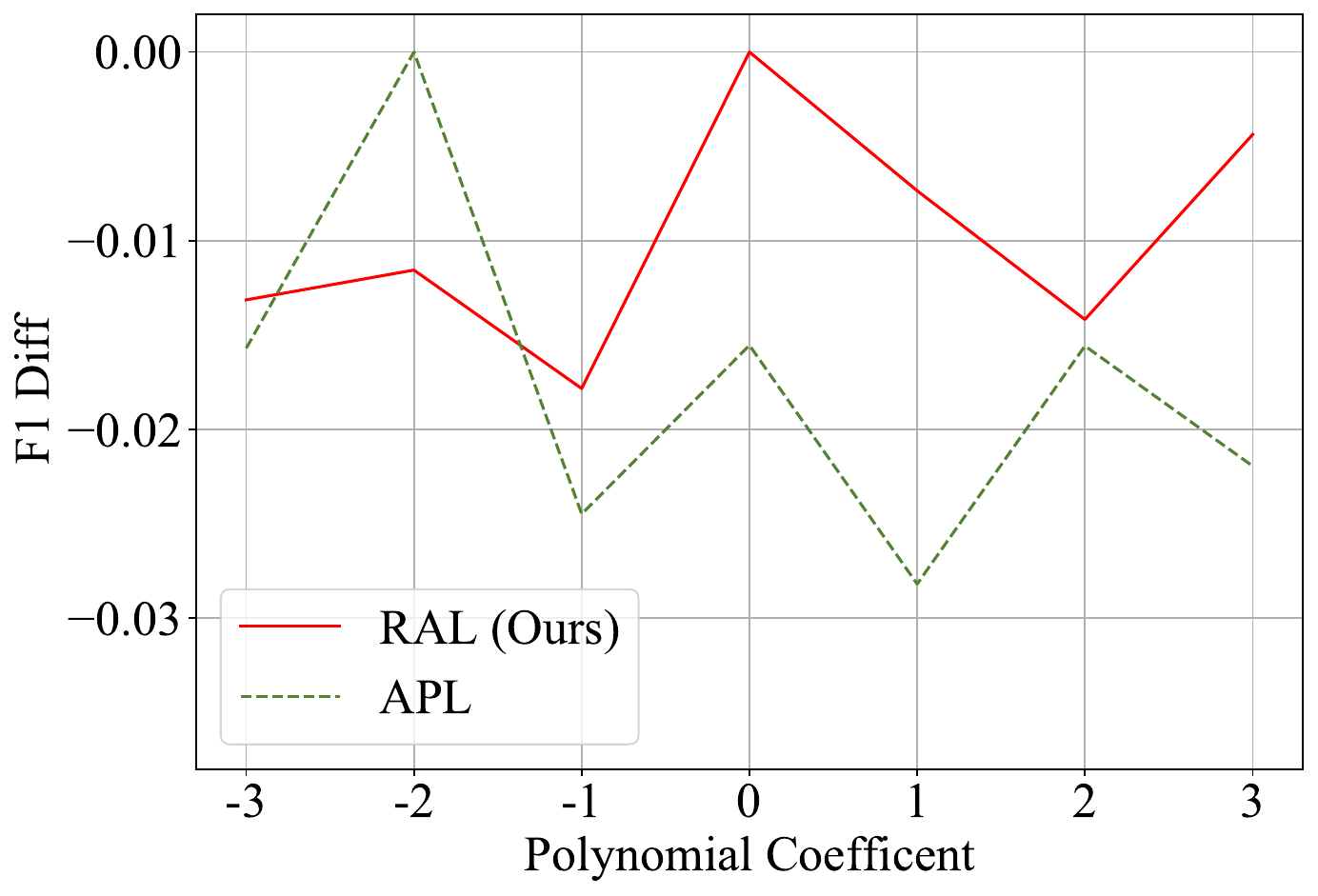}
    \caption{APTOS2019}
    \label{fig:apots-pow}
  \end{subfigure}
    \caption{
    Experimental result on the evaluation of polynomial coefficient. The Y-axis F1 Diff shows the difference from the best F1 score for each method (\ie APL and our RAL). Therefore, a value less than 0 indicates a larger difference from the best result and is sensitive to the polynomial coefficient hyper-parameter. In this result, \ours is less sensitive against the APL considerably.    
    }%
    \label{fig:pow_graph}%
\end{figure*}
\subsection{Robustness Analysis}
\begin{figure}[!t]
\centering
\includegraphics[width=.97\textwidth]{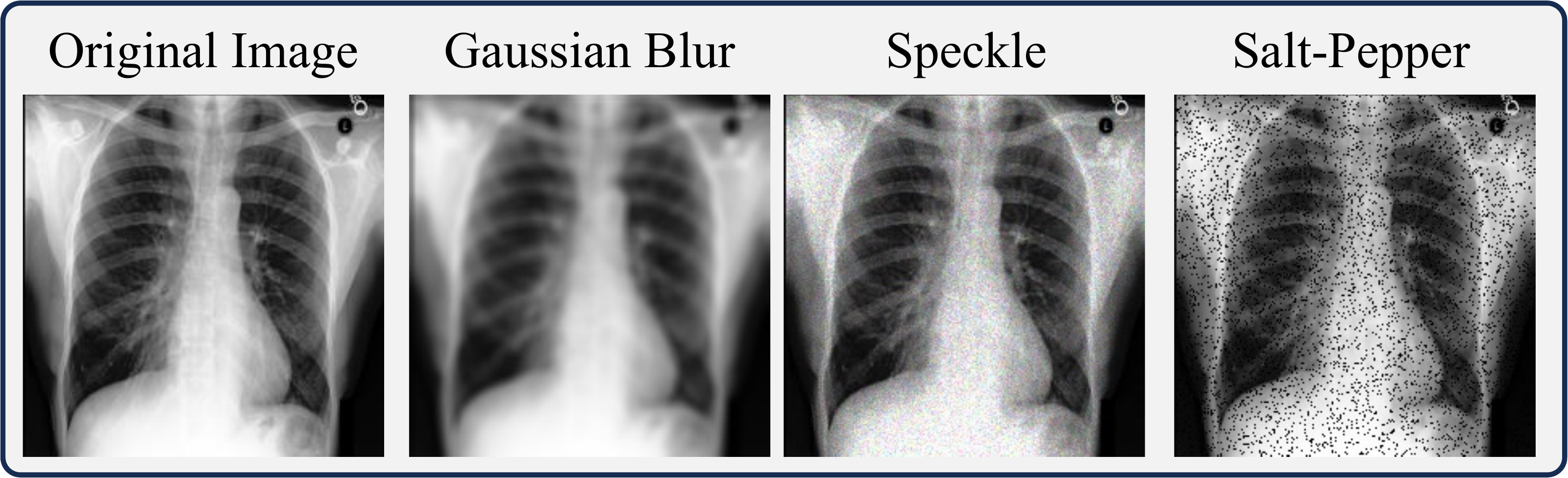}
\caption{Examples of noisy images such as Guassian Blur, Speckle, and Salt-Papper.}
\label{fig:noise_image}
\end{figure}

We carry out further experiments to investigate the robustness of \ours.
%
In our experiment, we introduce Gaussian Blur, Salt-Pepper, and Speckle noise to the original images of the CXR-LT dataset, as shown in Fig.\ref{fig:noise_image}.
%
To validate that our method performs well even with noisy conditions, we compare our method's mAUC to that of Binary Cross-Entropy(BCE) Loss and ASymmetric Loss(ASL).
%
Under the noisy condition, our method outperforms the others by about $1~3\%$ as shown in Table \ref{tab:noise_table}.
%
%
Therefore, our RAL has been empirically demonstrated to be more robust than existing methods when images are impacted by noisy conditions.

\begin{table}[!h]
    \centering
    \renewcommand{\arraystretch}{1.1}
    \begin{tabular}{ccccc}
    \toprule
        Methods & Img        & Blur  & Speckle  & SaltPepper               \\ \midrule
        BCE    & 0.789       & 0.718   &      0.502    &  0.501                        \\
        ASL    & 0.791      & 0.734  &     0.512     & 0.513                       \\
        RAL (Ours)   & 0.796  & 0.547  & 0.534 & 0.745         \\  \bottomrule
    \end{tabular}
    \caption{Experimental result on the noisy conditions. Our RAL shows better performance compared to the others consistently.}
    \label{tab:noise_table}
\end{table}

\subsection{Performance Analysis on Hyper-parameters}
We conduct more experiments for the performance analysis on hyper-parameter settings used in \ours. 
We compare ours with Asymmetric Polynomial Loss (APL)\cite{huang2023asymmetric} in this experiment.
We employ the same hyper-parameters for ours and APL.

First, we adjust the polynomial coefficient value of $\alpha_{m}$ in Eq.\ref{eq:RAL} for our RAL and APL. 
Figure \ref{fig:pow_graph} shows that, across all three datasets, our RAL is generally less sensitive to changes in the polynomial coefficient than APL, leading to less variance in the performance.
%

\begin{figure}[!ht]
  \begin{minipage}[b]{.52\linewidth}
    \includegraphics[width=\columnwidth, height=4.5cm]{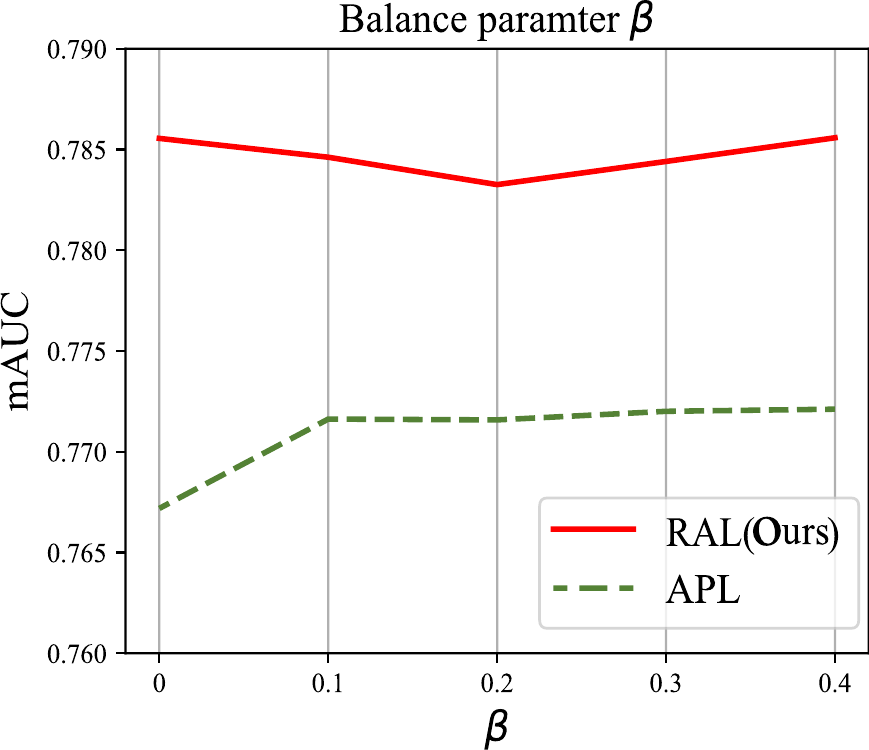}    
    \subcaption{Effect of parameter $\beta$.}
    \end{minipage} 
    \hfill
    \begin{minipage}[b]{.45\linewidth}
        \centering
        \begin{tabular}{@{}ccc@{}}
        \toprule
        $\lambda$ & F1 score  & mAUC \\ \midrule
        1.0       & 0.199 & 0.695    \\
        1.5       & 0.249 & 0.785    \\
        2.0       & 0.246 & 0.783    \\
        3.0       & 0.243 & 0.783    \\
        3.5       & 0.239 & 0.782    \\
        4.0       & 0.238 & 0.783    \\
        4.5       & 0.232 & 0.782    \\ \bottomrule
        \cr
        \end{tabular}%
        \subcaption{Effect of parameter $\lambda$.}
    \end{minipage}
\caption{Experimental result of ours according to the hyper-parameters $\beta$ and $\lambda$. 
In (a), we conduct an evaluation to compare our RAL with APL with regard to $\beta$ in Eq.\ref{eq:RAL}, the weight that regularizes the negative loss. In (b), we evaluate F1 and AUC scores in relation to $\lambda$ in Eq.\ref{eq:RAL}, which is utilized to avoid significant gradients in the negative loss.
}
\label{fig:hyper_test}
\end{figure}
    
Furthermore, we evaluate the balance parameter $\beta$, which governs the negative loss.
Fig.\ref{fig:hyper_test}(a) shows that for all the $\beta$ values, our RAL outperforms APL in terms of AUC score, highlighting the resilience to the hyper-parameter settings of our RAL.
In addition, we experimented with the balance parameter $\beta$, which controls the negative loss. 
%
Fig.\ref{fig:hyper_test}(b) shows the F1 and AUC score for different $\lambda$ values; it can be found that the best performance is obtained at $\lambda=1.5$ consistent with the previous study\cite{zhang2021simple}.

\begin{table}[!h]
    \centering
    \renewcommand{\arraystretch}{1.1}
    \begin{tabular}{cccc}
    \toprule
        Submit  &  mAP           & mAUC          & mF1   \\ \midrule
        \# 1    & 0.351         & 0.837         & 0.256  \\
        \# 2    & 0.317         & 0.814          & 0.061 \\
        \# 3    &  0.318        &  0.814         & 0.143  \\  \bottomrule
    \end{tabular}
    \caption{Experimental result of three submissions of test phase of the CVAMD 2023 competition.}
    \label{tab:Test_phase}
\end{table}

\section{Result on CVAMD2023 Competition}
Our method results in the Top-5 of the ICCV CVAMD 2023 competition's final rankings. To get this result, we scale the input image to $1024 \times 1024$ and use ConvNeXT-B models from \cite{liu2022convnet}. 
We increase the input image size in the test phase using the checkpoint file saved during the development phase.
We configure hyper-parameters in the same values as Sec.\ref{Imp_details}. 
In the final score of the competition, ours recorded 0.351 mAP, 0.837 mAUC, and 0.256 mF1 scores, which are included in the Top-5 ranking. With the efficient loss function design, we show improved performance on the multi-label long-tailed classification of the CVAMD 2023 challenge, which does not use additional model parameters or inference complexity. 
Table \ref{tab:Test_phase} shows the test phase results of our three submissions.

\section{Conclusion}
In this paper, we introduce the Robust Asymmetric Loss (RAL) for long-tailed multi-label classification tasks on medical images. Our proposed RAL trains the model more robustly against the various hyper-parameters without additional resources. RAL shows competitive results on the long-tailed single- and multi-label datasets compared to previous state-of-the-art loss functions. We especially achieve a Top-5 ranking in the CVAMD 2023 competition using our method. We think that future research can benefit from our findings and incorporate ours into their work.






{\small
\bibliographystyle{ieee_fullname}
\bibliography{egbib}
}

\end{document}